\documentclass[review]{elsarticle}

\usepackage{lineno,hyperref}
\usepackage{graphicx}
\usepackage{subfigure}
\usepackage{mathtools}
\usepackage{amssymb}
\usepackage{amsmath}
\usepackage{multirow}

\modulolinenumbers[5]

\journal{Visual Informatics}

\begin{document}

\begin{frontmatter}

\title{A Co-analysis Framework for Exploring Multivariate Scientific Data}
\tnotetext[mytitlenote]{
Received 30 November 2018, Accepted 22 December 2018, Available online 26 December 2018.
}
\tnotetext[mytitlenote]{
Fully paper is available on {https://www.sciencedirect.com/science/article/pii/S2468502X18300597}}

\author[]{Xiangyang He}

\author[]{Yubo Tao\fnref{*}}
\fntext[*]{Email: taoyubo@cad.zju.edu.cn (Corresponding author)}

\author[]{Qirui Wang}


\author[]{Hai Lin}
\address{State Key Lab of CAD\&CG, Zhejiang University}




\begin{abstract}
In complex multivariate data sets, different features usually include diverse associations with different variables, and different variables are associated within different regions. Therefore, exploring the associations between variables and voxels locally becomes necessary to better understand the underlying phenomena. In this paper, we propose a co-analysis framework based on biclusters, which are two subsets of variables and voxels with close scalar-value relationships, to guide the process of visually exploring multivariate data. We first automatically extract all meaningful biclusters, each of which only contains voxels with a similar scalar-value pattern over a subset of variables. These biclusters are organized according to their variable sets, and biclusters in each variable set are further grouped by a similarity metric to reduce redundancy and support diversity during visual exploration. Biclusters are visually represented in coordinated views to facilitate interactive exploration of multivariate data based on the similarity between biclusters and the correlation of scalar values with different variables. Experiments on several representative multivariate scientific data sets demonstrate the effectiveness of our framework in exploring local relationships among variables, biclusters and scalar values in the data.
\end{abstract}

\begin{keyword}
\texttt Multivariate data \sep bicluster \sep local association
\end{keyword}

\end{frontmatter}


\section{Introduction}
\label{section1}
Scientific simulations often generate data sets with multiple variables for complex physical phenomena. These variables generally include hidden associations, because of their collective application in the simulation model~\cite{Carr:2013:JCN}. For instance, a hurricane is a rapidly rotating storm system characterized by a low-pressure center, strong winds, including with heavy rains in climate simulation. However, the heterogeneity and complexity of multivariate data make the extraction of interesting associations, which are typically located only in the subspaces of variables and subsets of voxels, quite challenging. For example, the hurricane eye may possess a strong association with pressure and wind, while the eyewall clouds may be strongly associated with the water vapor and cloud moisture~\cite{Liu:2016:AAF}. Therefore, it would be meaningful to extract hidden associations between variables locally and detect local features based on the associated variables.

In multivariate data analysis, a broad variety of techniques have been proposed to explore the associations of variables/voxels in the data. At the voxel level, many clustering algorithms in the data mining field have been used to automatically group associated voxels as features~\cite{Tzeng:2004:ACV, VanLong:2009:MIV, Wu:2015:EAF}. Similarity metrics between voxels are generally defined as scalar values of all variables. Therefore, it may be challenging to detect features that only depend on a subset of variables because other unrelated variables may have a negative impact on clustering owing to the curse of dimensionality. At the variable level, many correlation metrics between variables such as gradient similarity measure (GSIM)~\cite{Sauber:2006:MAA}, Pearson product-moment correlation coefficients~\cite{Sukharev:2009:CSO}, and mutual information~\cite{Wang:2011:AIT} have been recently proposed. These metrics usually the average correlation values of all voxels. Because a subset of variables may be strongly associated in a local region, it is desirable to extract these local associations among variables in different local regions rather than the global associations based on all voxels. Furthermore, these methods automatically analyze the similarity of voxels or the correlation of variables independently. Variables and voxels should be analyzed together rather than separately to determine the local associations between them.

Multi-dimensional transfer functions can consider both variables and voxels in the manual classification of features of interest. For example, a feature can be specified by gradually selecting the scalar value intervals of a few variables in the parallel coordinate~\cite{Zhao:2010:MRT, Guo:2011:MTF} or by specifying Gaussian functions in a scatter plot matrix~\cite{Lu:2017:MVD}. In this case, these variables may be associated and their correlated scalar value intervals include the definition of the feature, the voxels of which include a similar scalar-value pattern over these associated variables. In this paper, we call it a bicluster between variables and voxels, which denotes two subsets of variables and voxels with close scalar-value relationships. While manual specification exhibits the advantage of flexibility in obtaining a variety of biclusters, it can be laborious and hinder comprehensive coverage of the data in exploratory analysis. In addition, this specification heavily depends on the domain knowledge and skills of users to determine the associated variables and their correlated scalar values in a meaningful feature. In practice, users must interactively refine the specification to search for satisfactory results. When there are many variables in multivariate data, it becomes nearly impossible to find all meaningful features because of the large size of the search space. This introduces the need to automatically find all meaningful biclusters between variables and voxels.

To address these requirements, we propose a co-analysis framework based on biclusters for interactively exploring multivariate data, including bicluster generation, analysis and visual exploration. Our framework first generates all biclusters (local relationships) by simultaneously clustering variables and voxels (Sec. \ref{sec:bicluster_generation}). The scalar values of voxels demonstrate a similar pattern on the variables in a bicluster, including a specific value combination. Because each bicluster is associated with a subset of variables, they can be organized by the variable set to hierarchically explore variables and biclusters. As some of the biclusters may overlap with each other because of the completeness of bicluster generation, biclusters with the same variable set can be grouped based on a similarity metric between biclusters (Sec. \ref{sec:bicluster_analysis}). To visually explore biclusters, we design a visual analysis system with four coordinated views to reveal three-faceted relationships in multivariate data (Sec. \ref{sec:bicluster_exploration}): variables, biclusters and scalar values. An association matrix between variables and biclusters is designed to facilitate searching for correlated variables. The biclusters in a variable set are displayed through dimensionality reduction to analyze the similarity of biclusters. An enhanced parallel coordinate plot is used to explore the correlation of scalar values of a group of biclusters or a bicluster. Based on the exploration guideline, we experiment on several multivariate data sets in different domains to demonstrate the effectiveness and usefulness of our co-analysis framework (Sec. \ref{sec:bicluster_exploration}).

This paper is an extended version of our conference paper~\cite{He:2018:BBV} of \emph{IEEE Scientific Visualization Conference (SciVis)}. In detail, the extensions of our conference paper include the following:
\begin{itemize}
  \item {a detailed description of bicluster generation based on the variance minimization method (Sec. \ref{sec:bicluster_generation}),}
  \item {a new correlation measure between variables based on biclusters (Sec. \ref{subsec:bicluster_organization}),}
  \item {an enhanced parallel coordinate plot for presenting the statistical distribution of associated voxels (Sec. \ref{subsec:scalar_value_view}), and, }
  \item {two new experiments on the ionization front instability data (Sec. \ref{subsec:association_matrix}) and hurricane Isabel data (Sec. \ref{subsec:hurrican_isabel}).}
\end{itemize}


\section{Related Work}

Multivariate data analysis and visualization, as one of the major challenges associated with scientific visualization, have long been active research topics~\cite{He:MSD, Kehrer:2013:VVA, Fuchs:2009:VMC}. In this section, we briefly review previous research related to correlation analysis, and interactive classification of multivariate data.

\subsection{Correlation Analysis}

Finding hidden correlations in multivariate data is a common challenge in many computational analysis fields. Many correlation analysis methods have been proposed over the years to explore the relationships between variables and scalar values.

Information theory provides a theoretical framework of measuring the global correlation between variables. Biswas et al.~\cite{Biswas:2013:AIF} employed mutual information to measure the informativeness of one variable about the other variable and grouped variables based on mutual information in a graph-based approach. Wang et al.~\cite{Wang:2011:AIT} applied transfer entropy to investigate the causal relationships between variables in time-varying multivariate data, and the correlations between variables were visually encoded in a node-link diagram. These methods consider the entire data, making it challenging to capture local correlations between variables in different regions.

Many local correlation metrics have been proposed to capture the correlation at each voxel, and the correlation between variables can be measured by the summation of the correlation values of all voxels. Sauber et al.~\cite{Sauber:2006:MAA} proposed a gradient similarity measure (GSIM) and a local correlation coefficient to measure the local correlation at each voxel. In addition, they introduced a multifield-graph to present an overview of the correlation between variables.
Gosink et al.~\cite{Gosink:2007:VII} derived a correlation field by taking the normalized dot product between two gradient fields from two variables. Janicke et al.~\cite{Jnicke:2007:MVU} extended the concept of local statistical complexity to multi-fields to identify spatiotemporal structures that exhibit the same behavior in multivariate data. Sukharev et al.~\cite{Sukharev:2009:CSO} applied the Pearson product-moment correlation coefficient on temporal curves of voxels to analyze the linear correlation between two variables in time-varying multivariate data. Nagaraj et al.~\cite{Nagaraj:2011:AGC} presented a gradient-based correlation criterion, the norm of a partial derivative matrix, to capture the interactions between multiple scalar fields. The correlation field is visualized to detect regions with high correlation values. In this study, we simultaneously cluster variables and voxels to automatically extract biclusters and employ the biclusters, subsets of voxels instead of all voxels as in previous methods, to better measure the correlation of a subset of variables in local regions.

In addition to the correlation between variables, a specific local relationship between the scalar values in different variables has recently received considerable attention. Biswas et al.~\cite{Biswas:2013:AIF} applied the surprise and predictability metrics to measure the variability of one scalar value with respect to another variable. Liu et al.~\cite{Liu:2016:AAF} considered two-way interactions between the scalar values of two variables as information flow and employed the association rules to model these interactions. Because one bicluster includes voxels with a similar scalar-value pattern over a subset of variables, we can directly analyze such local relationships between scalar values in multiple variables.



\subsection{Interactive Classification}

Feature classification is essential for effective exploration of multivariate data. For multivariate non-spatial data, there are many well-studied visualization and interactive exploration techniques that display the distribution and the relationship of data. These techniques include parallel coordinates and scatter plots. Parallel coordinates illustrate information on each dimension including the correlation between neighborhood axes~\cite{Inselberg:1985:TPW, Inselberg:1990:PCA}. Scatter plots display high-dimensional data using dimensionality reduction techniques, such as multidimensional scaling (MDS)~\cite{Cox:1994:MDS} and t-SNE~\cite{Maaten:2008:VDT} because it can be easy to identify and select clusters after projection. Previous classification methods of multivariate data primarily rely on these techniques.

In multivariate data visualization, Zhao and Kaufman~\cite{Zhao:2010:MRT} introduced parallel coordinates for multi-dimensional transfer function design, which defines a feature by specifying several scalar value intervals of associated variables. Guo et al.~\cite{Guo:2011:MTF} proposed a novel transfer function design interface combining the parallel coordinate and MDS plots to facilitate feature specification in multivariate data. Lu and Shen~\cite{Lu:2017:MVD} presented a bottom-up subspace exploration workflow that allows users to interactively design multivariate transfer functions, and introduced additional information that guides users in the selection of subspaces, discovering interesting features. While multi-dimensional transfer functions flexibly enable specific features, it can be time consuming and challenging to search for all meaningful features in the exploratory analysis. In this study, we automatically extract all meaningful biclusters, and visually explore the similarities of biclusters in the scatter plot, including the correlation of scalar values in a bicluster in the parallel coordinate.

\section{Overview}

\begin{figure*}[]
 \centering
 \includegraphics[width=0.97\textwidth]{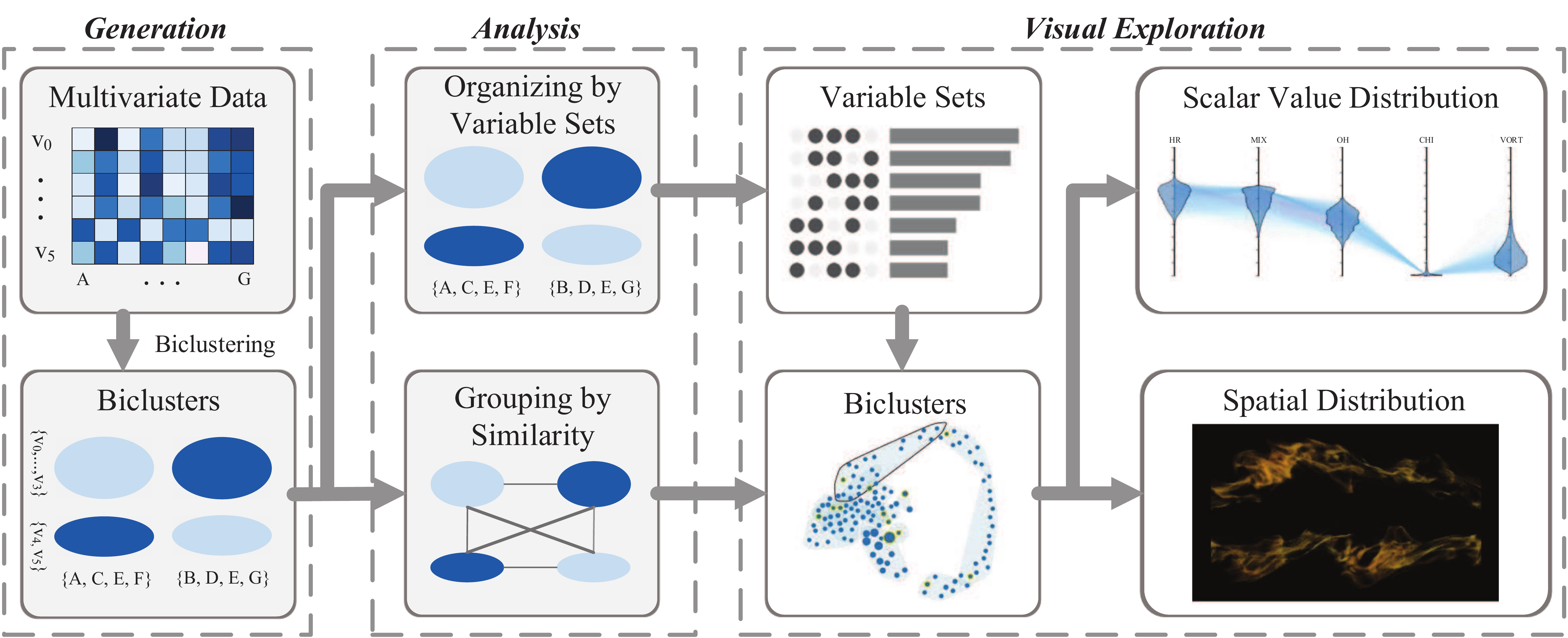}
 \caption{The co-analysis framework for multivariate data. This example includes eight variables, A-G, and six voxels, v$_0$-v$_5$. Our framework first generates all biclusters by simultaneously analyzing variables and voxels. These biclusters are organized by their variable sets and are hierarchically grouped based on a similarity metric in the analysis stage. Four coordinated views are designed to visually explore the local relationships in variables, biclusters, and scalar values.}
 \label{fig:framework}
\end{figure*}

Biclustering, also known as co-clustering or simultaneously clustering~\cite{Hartigan:1972:DCO}, addresses this problem by simultaneously clustering both rows (objects) and columns (attributes) in a variety of fields such as DNA micro-array data analysis, text mining, and information retrieval. Biclustering has also been widely used in visualization, such as visualizing relevant genes and conditions of gene-expression matrices~\cite{santamaria:2008:bicoverlapper}, reducing visual clutter of a large number of edges~\cite{Liu:2017:TBA}, interpreting the subspace clustering result in high-dimensional data~\cite{liu:2015:visual}, and discovering coordinated relations from textual datasets~\cite{Sun:2016:BSE}. In the field of data mining, the biclustering method can effectively extract cohesive objects with a similar scalar-value pattern over a subset of attributes. This study applies biclustering to the analysis of multivariate scientific data. If variables and voxels are considered attributes and objects respectively, we can extract cohesive voxels with a similar scalar-value pattern over a subset of variables using the biclustering method. Therefore, a bicluster is composed of a subspace of variables and a subset of voxels, and these voxels demonstrate a similar scalar-value pattern on these variables, including a specific value combination of these variables. In other words, these variables are locally associated in the space of these voxels, and the corresponding scalar values of these variables are strongly correlated with each other. In this way, a bicluster provides a local association of variables and scalar values in these voxels. Furthermore, the corresponding spatial region of a bicluster can be viewed as a feature of multivariate data.

Based on the concept of biclusters, our co-analysis framework, as shown in Fig.~\ref{fig:framework}, includes bicluster generation, analysis, and visual exploration to guide users in the exploration of various facets of local relationships in multivariate data. We first automatically extract all biclusters from multivariate data based on a biclustering method. Each bicluster contains a local relationship among variables, voxels, and scalar values.

Because the biclustering method is used to generate all possible biclusters, the number of biclusters can be very large, and some of them may overlap with each other. It becomes nearly impossible for users to interactively analyze these biclusters one by one. Therefore, we first hierarchically organize biclusters based on their variable sets, and users can explore biclusters in the context of a variable combination. We then design a similarity metric between biclusters to group them to reduce redundancy. This corresponds to a semantic analysis of the three-level analytical tasks of biclusters~\cite{Zhao:2018:BVE}.


For visual analysis of biclusters, we propose a visual analytics system with four coordinated views to interactively analyze local relationships in multivariate data, including the correlation of the variable set, the similarity of biclusters, the correlation of scalar values in biclusters, and the spatial distribution of biclusters. All these views are linked to support the interactive exploration of local relationships in variables, biclusters, and scalar values.

\section{Bicluster Generation}
\label{sec:bicluster_generation}
There are many categories of biclustering/co-clustering methods, which can be used to generate biclusters. The main difference between them is the clustering strategy. One of the most popular biclustering methods is the variance minimization method~\cite{Oghabian:2014:BM}, which has been extensively studied under the name of pattern-based clustering. The basic assumption is that an object often exhibits a similar scalar value pattern over several attributes. Therefore, this study employs the variance minimization method to generate biclusters, because a local feature/phenomenon in multivariate data may also illustrate similar scalar-value patterns over several variables, i.e., a specific value combination of a subset of variables, including two associated scalar values in two variables~\cite{Liu:2016:AAF}.

\begin{figure}
\centering
\subfigure[]{\includegraphics[width=0.32\textwidth]{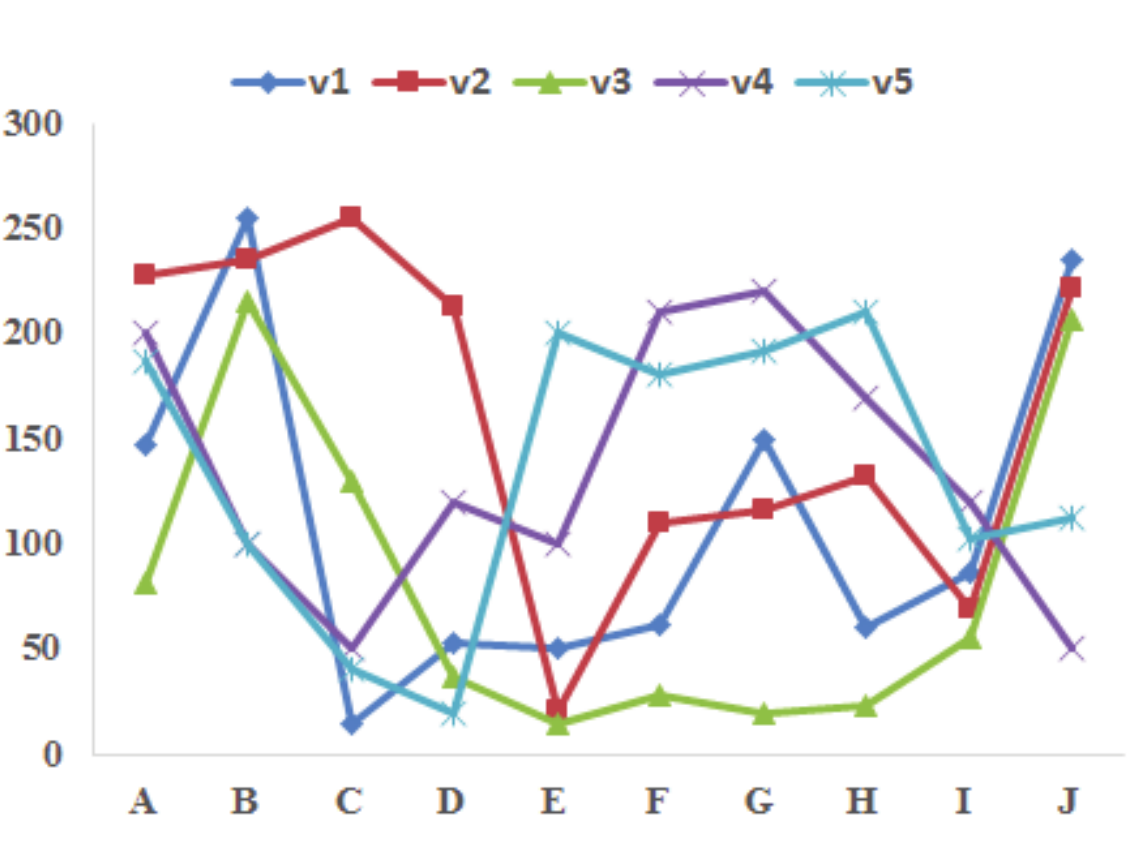}\label{subspace-all}}
\subfigure[]{\includegraphics[width=0.32\textwidth]{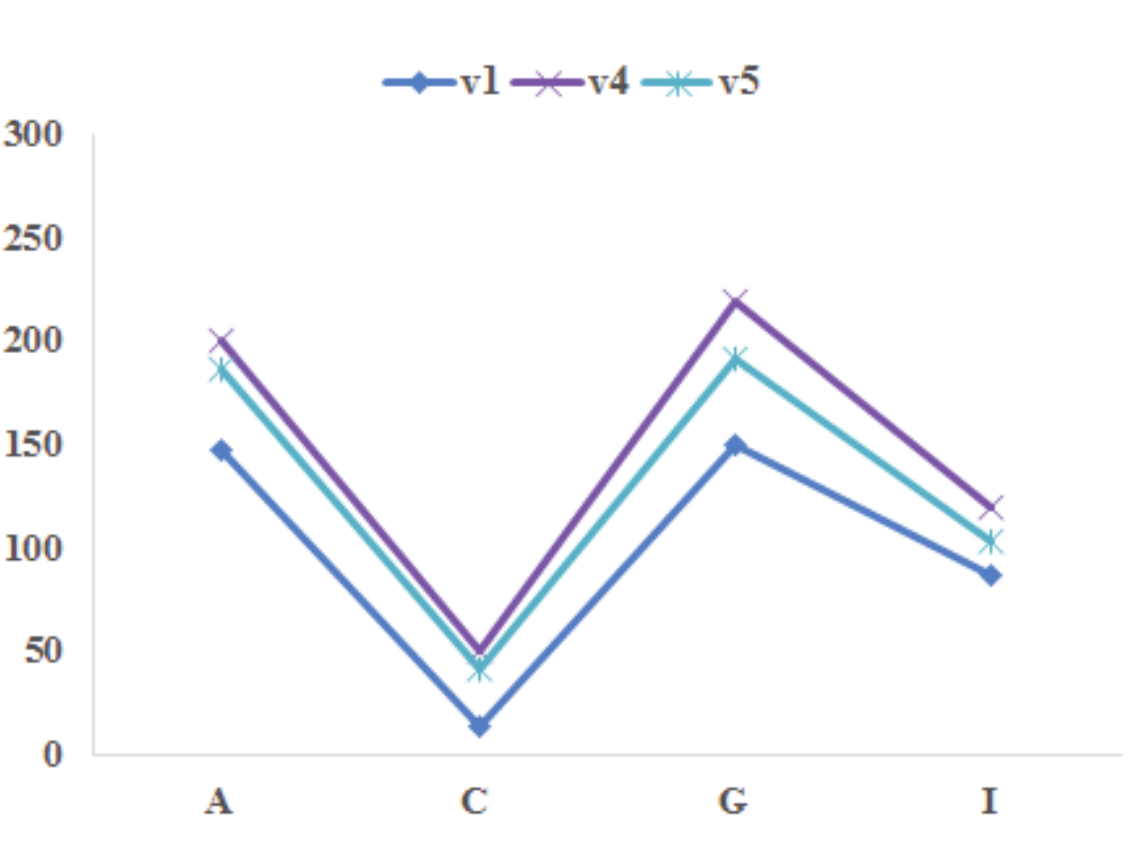}\label{subspace-one}}
\subfigure[]{\includegraphics[width=0.32\textwidth]{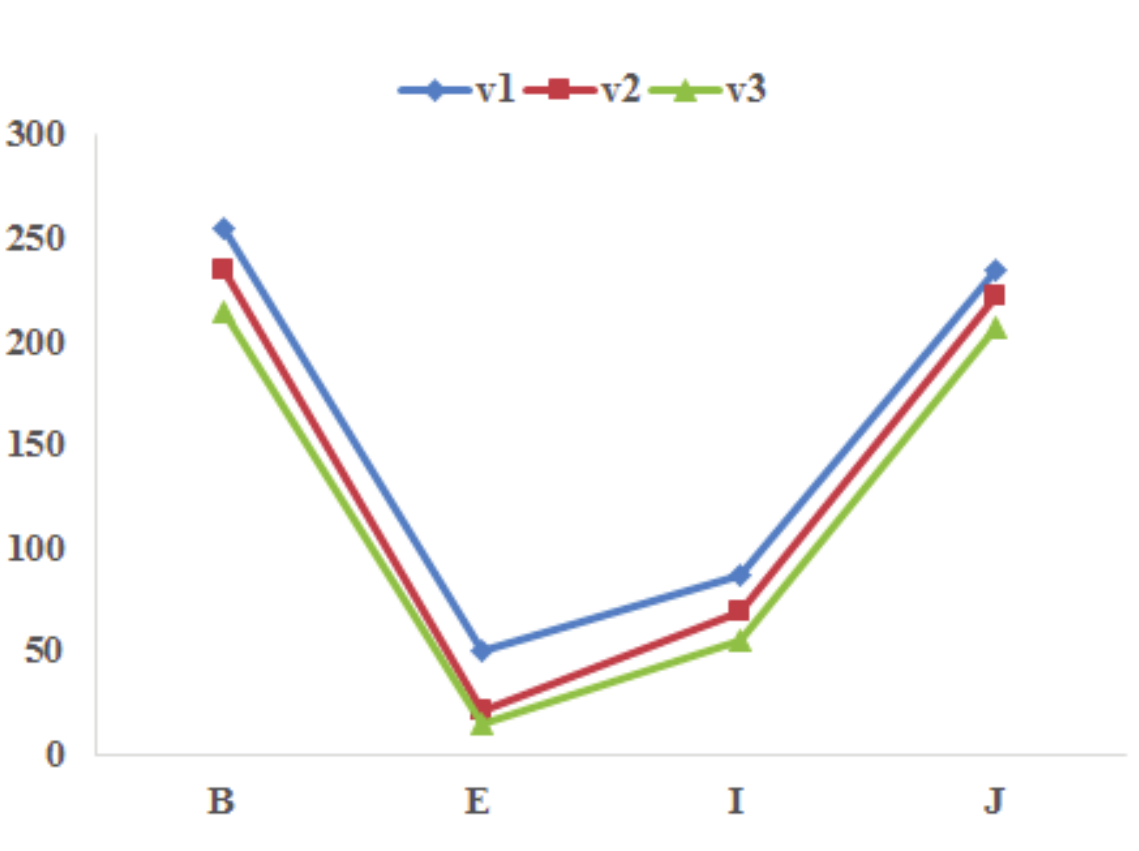}\label{subspace-two}}
\caption{An illustration of biclusters. (a) A data set with 10 variables and 5 voxels shows no clear patterns. (b) The bicluster (\{A, C, G, I\}, \{$v_1, v_4, v_5$\}) reveals a clear pattern. (c) The bicluster (\{B, E, I, J\}, \{$v_1, v_2, v_3$\} shows another clear pattern.}
\label{fig:subspace}
\end{figure}

For example, let us consider ten variables with five voxels in Fig.~\ref{subspace-all}. It is obvious that there is no clear pattern among five voxels. However, if we choose two subsets of variables as shown in Fig.~\ref{subspace-one} and~\ref{subspace-two}, respectively, the voxels $v_1, v_4$ and $v_5$ follow a similar scalar-value pattern, i.e., a coherent pattern, in the variable set \{A, C, G, I\}, while the voxels $v_1, v_2$ and $v_3$ share another coherent pattern in the variable set \{B, E, I, J\}. The variance minimization method is an effective method for extracting these pattern-based biclusters automatically through the simultaneously analysis of variables and voxels. This can lower the requirement of the domain knowledge and skills of users when performing explorative analysis of multivariate data. Because the phenomena in the data may be not well separated, biclusters do not need to be exclusive. Therefore, this study chooses Maple~\cite{Pei:2003:MAF}, an important algorithm in the variance minimization method as the basis of the co-analysis framework because it can identify overlapped clusters and guarantee completeness of the bicluster search.

We first organize all variables (dimensions) $D=\{d_0, d_1, ... , d_{M-1}\}$ and voxels $V=\{v_0, v_1, ... , v_{N-1}\}$ in a variable-voxel matrix, where $M$ and $N$ are the number of variables and voxels in multivariate data, respectively. Each entry $s_{i,j}$ in the matrix is the scalar value of the $j$-th variable at the $i$-th voxel. The coherent of two voxels $v_x$ and $v_y$ on two variables $d_u$ and $d_v$ is measured by the pScore~\cite{Pei:2003:MAF} as follows:
\begin{equation}
pScore(\begin{bmatrix}
s_{x,u} & s_{x,v} \\
s_{y,u} & s_{y,v}
\end{bmatrix})=\left \| ({s_{x,u}}-{s_{x,v}})-({s_{y,u}}-{s_{y,v}}) \right \|.
\end{equation}
The pScore restricts the coherence to a $2\times2$ matrix and describes the change in scalar values on two variables between two voxels. Clearly, the smaller the pScore, the greater the coherence of the two voxels on two variables. pScore is more rigorous than other pattern definitions and more robust to noises~\cite{Kriegel:2009:CHD}.

Using the pScore, we can define a bicluster $(D', V')$, $D' \subset D$ and $V' \subset V$, if the pScore of any two voxels $v_x, v_y \in V'$ on any two variables $d_u, d_v \in D'$ is less than or equal to a user-specified tolerance $\delta$. Most biclusters correspond to specific value combinations of voxels in $V'$ on variables in $D'$ within a tolerance, such as the biclusters in Fig.~\ref{fig:subspace}. Scientists are more interested in specific value combinations of multi-variables to obtain depth knowledge about the interaction of variables in simulations. Previous methods usually apply data binning to enforce the tolerance before searching for specific value combinations~\cite{Liu:2016:AAF}, while we apply the tolerance during the clustering process to generate more complete results. A bicluster is closed if adding any voxels violates the above definition. Therefore, we only need to consider closed biclusters in one variable set, and we refer to closed biclusters as biclusters for simplicity.

\begin{figure*}
 \centering
 \includegraphics[width=0.97\textwidth]{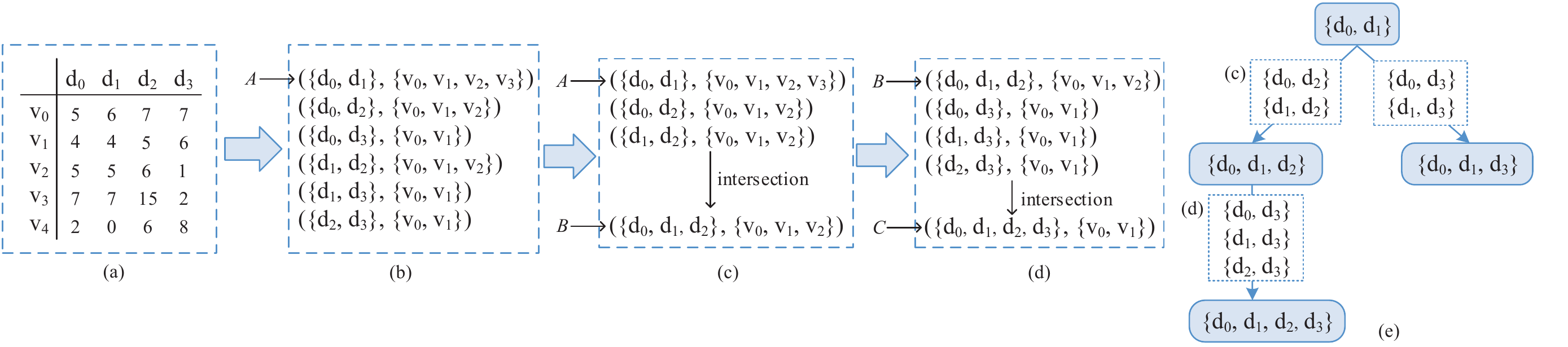}
 \caption{A running example of bicluster generation. (a) A data matrix with four variables and five voxels. (b-d) The first step is to generate biclusters for all pairs of variables, then the bicluster \emph{A} is expanded to a new biclusters \emph{B} and \emph{C} successively. (e) A variable enumeration tree illustrates the depth-first searching process for the bicluster \emph{A}. The left subtree corresponds to the searching process (c) and (d), and the searching process of the right subtree is finished after adding the variable $d_3$, as the variable set $\{d_0, d_1, d_2, d_3\}$ has been explored in the left subtree.}
 \label{fig:algorithm}
\end{figure*}


A depth-first searching algorithm is first used to find biclusters in lower dimensions, and the biclusters are then merged to derive biclusters of higher dimensions. The variable set is iteratively expanded from two variables to all variables in the manner of multiple trees. If an expanded variable set has been previously explored in the same tree, the searching process can be stopped for this subtree to improve computational efficiency as its bicluster and its children have already been generated.

An example of the searching process with $\delta$ = 1 is illustrated in Fig.~\ref{fig:algorithm}. We systematically enumerate every combination of variables using a variable enumeration tree and a depth-first search. As shown in Fig.~\ref{fig:algorithm}(b-d), biclusters in two variables are first generated based on the variable-voxel matrix, and then the variable set extends one by one until the number of variables reaches the maximum value. The voxel set of the extended variable set is generated by the intersection of corresponding voxel sets. We repeat this searching process for each bicluster to generate all biclusters using the depth-first searching process illustrated by the variable enumeration tree in Fig.~\ref{fig:algorithm}(e), and the enumeration procedure is similar to mining frequent closed itemsets.

In multivariate data, the number of voxels is significantly large than the number of variables. Each variable set includes multiple associated voxel sets, i.e., multiple biclusters with the same variable set. In the exploration of multivariate data, a bicluster may be statistically insignificant if it contains a small number of voxels, and this can also reduce the searching time of biclusters. Therefore, there are two parameters in the depth-first searching algorithm: $\delta$, the tolerance ($pScore(D', V')\leqslant \delta$), and $minv$, the minimal number of voxels ($\left | V' \right | \geqslant minv$).

\section{Bicluster Analysis}
\label{sec:bicluster_analysis}
Because the biclustering method guarantees the completeness of the bicluster search, we acquire all biclusters in multivariate data. The generated biclusters are not necessarily exclusive, which means that a voxel/variable can appear in more than one bicluster. Consequently, the number of biclusters is generally very large, and some of them are very similar. To facilitate visual exploration of biclusters, it is necessary to organize and group biclusters to reduce redundancy and encourage diversity in during visual exploration.

\subsection{Bicluster Organization}
\label{subsec:bicluster_organization}
Each bicluster is associated with one variable set, and one variable set is generally associated with multiple biclusters. Because the number of variable sets, i.e., the variable combination, is much smaller than the number of biclusters, we first organize biclusters based on their variable sets. The variable sets can be organized hierarchically, and we can iteratively expand the variable set from two variables to multiple variables to reduce the complexity of the bicluster analysis.

While previous methods measure the global correlation of multiple variables~\cite{Sukharev:2009:CSO, Wang:2011:AIT}, we can measure the local correlation of multiple variables in a variable set by analyzing its associated biclusters. Because the scalar values of voxels are generally linear between two variables in a bicluster, we choose the Pearson correlation coefficient to evaluate the linear correlation between two variables. As the voxels have a similar scalar-value pattern over these variables in the variable set, it would be better to use only associated voxels to measure the local correlation of multiple variables in the variable set, instead of all voxels in previous methods. The correlation of multiple variables is the minimal absolute value of the Pearson correlation coefficient of each pair of variables in the variable set $D'$, as follows:
\begin{equation}
C(D') = min\{ {\left | \frac{cov(d_u, d_v)}{\sigma_{d_u}\sigma_{d_v}} \right |} \colon d_u, d_v \in D'\},
\end{equation}
where $cov$ is the covariance, and $\sigma_{d_u}/\sigma_{d_v}$ is the standard deviation of the variable $d_u/d_v$ in the voxels of biclusters of the variable set. The correlations of variable sets can help users choose the variable set to explore first.



\subsection{Bicluster Grouping}
\label{sec:bicluster_grouping}
Some of the biclusters may overlap with each other, especially biclusters with the same variable set. Therefore, we hierarchically group biclusters with the same variable set to yield a smaller set of mutually sufficiently different, yet individually interesting groups of biclusters for interactive exploration.

Grouping quality primarily depends on the similarity metric between two biclusters. Because biclusters to be grouped, have the same variable set, and there is one to one mapping between voxels and scalar values for each variable, the similarity metric must only consider voxels in the bicluster. One promising similarity metric is the spatial overlap because the spatial distribution is a more intuitive way to recognize a feature in volume visualization. If two biclusters have more common voxels, i.e., a large spatial overlap, they are more similar to each other. Therefore, the similarity metric is defined as the Jaccard similarity coefficient as follows:
\begin{equation}
J(A, B) = \frac{\left | V_{A}\bigcap  V_{B} \right |}{\left | V_{A}\cup  V_{B} \right |},
\end{equation}
where $V_A$ and $V_B$ are the voxels of two biclusters $A$ and $B$, respectively.


Using the similarity metric, the agglomerative hierarchical clustering~\cite{Han:2011:DM} is applied to group the biclusters. The distance between two biclusters $A$ and $B$ is defined as $d(A, B) = 1 - J(A, B)$. When combining two groups of biclusters, a weighted average linkage criterion, a recursive definition for the distance between two groups, is used to compute the distance between two groups. For each group, one representative bicluster, such as the one with the largest number of voxels, is selected to guide users in the exploration of large or unfamiliar biclusters in multivariate data. Note that the similarity metric and clustering method is a choice of currently available methods that work for multivariate data. It could be replaced by other similarity metrics that are more suited to the specific requirements of data.

\section{Bicluster Exploration}
\label{sec:bicluster_exploration}
Using feature subspaces, including their groups and clusters, we design four coordinated views to visually identify, interpret and compare the local relationships in multivariate data.

\subsection{Association Matrix}
\label{subsec:association_matrix}
\begin{figure}
\centering
\includegraphics[width=0.7\columnwidth]{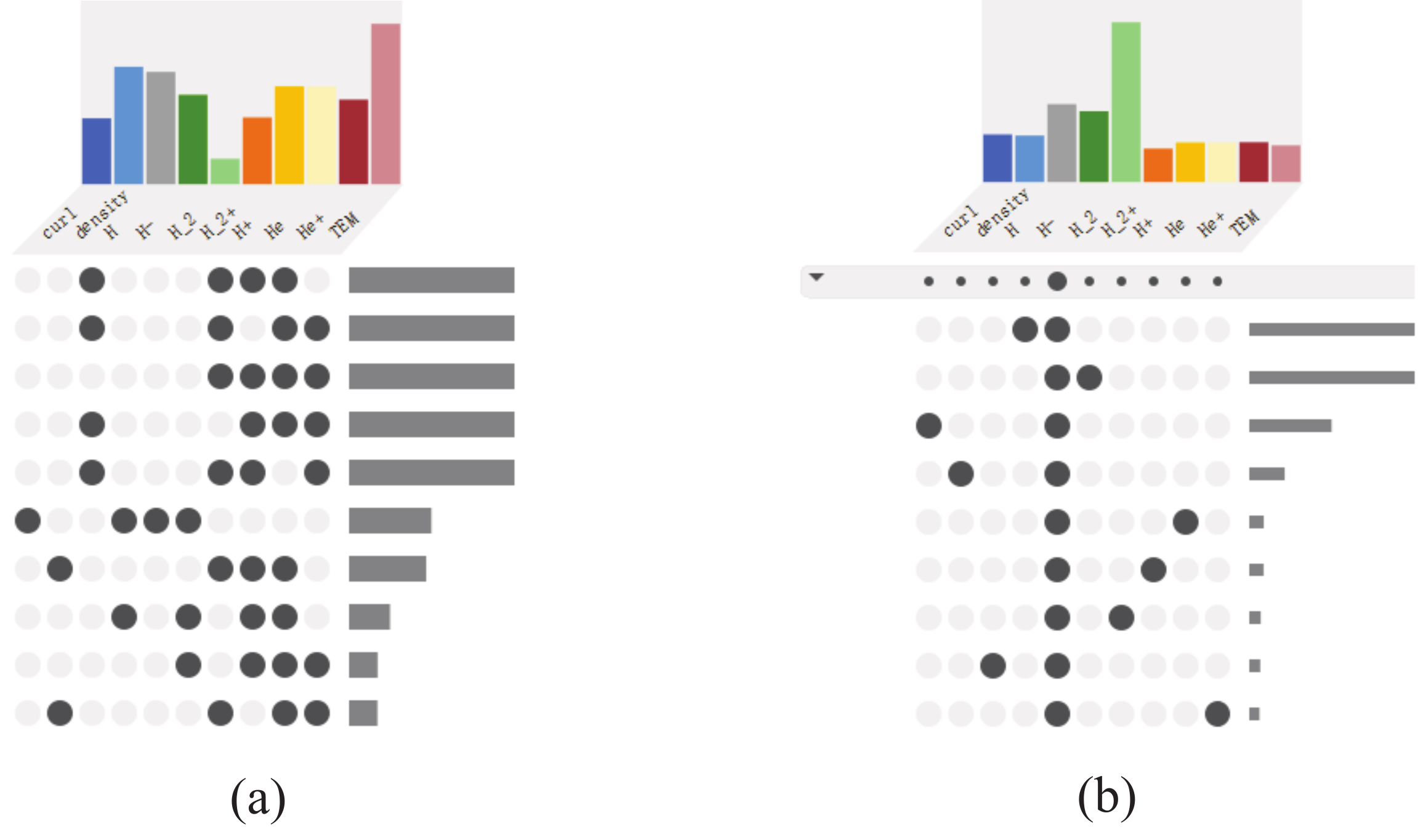}
\caption{The association matrix for the ionization front instability data set. (a) The variable sets with at least four variables are sorted by their correlation values. (b) The variable H$_2$ is drilled down to its children variable sets, and the correlation between H$_2$ and other variables is shown in the top bar chart.}
\label{fig:matrix}
\end{figure}

The variable sets are inherently hierarchical because of the searching process in bicluster generation. The hierarchical structure of the variable sets is helpful for users to iteratively explore biclusters.
We propose an association matrix to display the hierarchical structure of variable sets. The matrix layout is inspired by the combination matrix for the quantitative analysis of sets, their intersections, and aggregates of intersections~\cite{Lex:2014:UVI}. Each column in the association matrix corresponds to a variable of multivariate data, and each row corresponds to a variable set, including the associated biclusters of the variable set. The rows without associated biclusters are hidden by default, but they can be shown on demand during visual exploration. The variable in the variable set is encoded with a filled dark circle, otherwise a light-gray circle, as shown in Fig.~\ref{fig:matrix}. Therefore, it is more intuitive to recognize the variables in the variable set in each row, and the names of variables are listed in the top of the matrix. Additional attributes of the variable set can be displayed via the bar chart on the right of each row, and the length of the bar is proportional to the value of the attribute. The matrix layout enables the effective representation of associated data and additional summary statistics, and provides an overview of the relationships between the variable and the bicluster.

As for all matrix-based techniques, sorting is important to ensure an efficient representation of the data. Therefore, we offer various sorting options to analyze the relationship among variables in the local regions and derive a meaningful variable set, i.e., biclusters, for detailed exploration. The sorting attributes primarily include the number of variables in the variable set (the cardinality), the correlation of the variable set, and the number of biclusters in the variable set. We can also use these attributes to filter out less interesting variable sets and reduce the exploration space. For example, the variable sets with at least four variables are sorted by the descending correlation value in Fig.~\ref{fig:matrix}(a). The first five variable sets have a high correlation value, and this can guide users to focus on these variable sets first.


We also support drilling down from one variable set to its children variable sets, which is similar to expanding a node in the radial tree, to hierarchically explore biclusters. The variables in the expanded variable sets are encoded by smaller dark circles, and other variables are encoded by dark points. The bars associated with expanded rows have a reduced width to distinguish different levels. These children variable sets can be sorted by another attribute for visual comparison. For example, if a user attempts to determine which chemical species are most similar to H$_2$, he/she can choose the variable H$_2$ as the starting variable. As shown in Fig.~\ref{fig:matrix}(b), sorted by the descending correlation value, it can be straightforward to obtain the most correlated chemical species H- and H$_2$+ (the first two rows).


By sorting and drilling down, it becomes easy for users to identify variable sets of interest for detailed analysis by choosing the top rows in the matrix. We additionally provide the correlation information between variables based on information theory. This information is calculated on all voxels, instead of the voxels in biclusters, and they are displayed in the bar chart on the top of the matrix. The bar chart illustrates the entropy of each variable by default, and displays mutual information when one variable is selected. Therefore, users can select one variable of interest based on their domain knowledge, and drill down to its children variable sets to choose a variable set for detailed exploration.


\subsection{Bicluster View}

When one variable set is chosen in the association matrix, we must analyze and compare its biclusters, especially the similarity between them. Dimensionality reduction methods have been widely used for similarity analysis in 2D. Therefore, we apply MDS~\cite{Guo:2012:SMV}, one of the widely used dimensionality reduction methods, to project the biclusters of the variable set based on the spatial overlap similarity in the bicluster view. The scatter plot provides an overview of the similarity between biclusters, as shown in Fig.~\ref{fig:isable}. Each cycle is a bicluster, and its size is proportional to the number of voxels in the bicluster.



Because biclusters are clustered hierarchically in Sec 5.2, approximately 10 groups are selected to illustrate the meaningful features and local correlations and simplify the interactive exploration. The groups must be coherent enough ($d(A, B) \leqslant 0.99$ in our experiments) and the number of groups must not be too large to avoid visual clutter and to help users in the selection and analysis of biclusters. Each group is encoded by a light-blue and convex region, which covers all biclusters in the group. The representative bicluster of each group is highlighted by orange halos to distinguish between them. Because of the projection error, the regions of groups may be overlapped and result in the confusion; i.e., it is unclear as to which group do the biclusters in the overlapped area belong to. Therefore, when hovering with the mouse over the region of one group, its biclusters are highlighted to demonstrate the membership. For a correlated variable set, there could be dozens or even hundreds of biclusters, which would result in visual clutter because of the limited projection space. In this case, only the representative biclusters could be displayed for groups, and other biclusters in one group can be shown on demand. Users can select one group or one bicluster to further explore its scalar-value and spatial distribution to identify local correlations and meaningful features.

As biclusters cannot be grouped perfectly, we allow the manual verification and modification of groups. It is difficult to decide from the bicluster view alone if a bicluster belongs to a group because the scalar-value and spatial distributions are more critical in the similarity analysis of biclusters.  Owing to hierarchical clustering, we can select one group by clicking on its region to verify its distribution both in the scalar value and space, and split the group (move the next level) into two groups if it is highly diverse in the distribution. Two groups can be merged into one group if they are very similar. Through these refinements, we can better understand the similarity of biclusters and interactively identify meaningful local correlations.

\subsection{Scalar-Value View}
\label{subsec:scalar_value_view}
When one group or bicluster is selected, we employ a parallel coordinate to display its scalar-value distribution over its variables in the scalar-value view, as shown in Fig.~\ref{fig:isable}. The axis of each variable in the variable set is moved to the front, or the axes of other variables are hidden to facilitate the correlation analysis between the scalar values and variables. The parallel coordinate usually draws the polylines on the axes directly, making it hard to interpret the density of scalar values. We enhance the parallel coordinate by counting the occurrences of each scalar-value pair between neighbor axes, and using this information, we encode the transparency of the color. Although the transparency makes the density distribution more observable, it is still not easy to visually compare the density for some cases. Therefore, the histogram of the scalar values is rendered on both sides of the axes to further enhance the parallel coordinate. This eases the analysis of the coherence of the scalar values in one variable and the identification of the range of scalar values that includes most voxels.

For a group, the parallel coordinate can be used to verify the similarity of biclusters in the group. If the scalar-value distributions on the axes are all within a small range, biclusters are similar to each other in the group. Otherwise, this group may be split to generate coherent groups. For a bicluster, the parallel coordinate can be used to present the coherent scalar-value pattern, and to analyze the manner in which the specific values of the variables interact with each other.

\subsection{Spatial View}

In addition to the scalar-value distribution of one group or bicluster, the spatial distribution is important for local correlation analysis. The probability of the voxels belonging to a group or bicluster is calculated. The probability volume is visualized by direct volume rendering to display the spatial distribution and analyze the spatial coherence of a group or bicluster.

\subsection{Exploration Guideline}

Our co-analysis framework provides an analysis guideline that enables users to explore various aspects of multivariate data as an overview or in detail, as illustrated in Fig.~\ref{fig:isable}. Given a multivariate data set, the variable set of interest can be obtained by sorting according to application-related attributes, such as the correlation of the variable set and the number of biclusters or the mutual information between variables in the bar chart. The variable set can be drilled down to identify the variable set most related to a feature/phenomenon in the association matrix. Using the selected variable set, the clusters and their biclusters are illustrated in the scatter plot, and presenting an overview of their similarities to user. Users can visually explore each cluster or representative bicluster, and analyze the association of scalar values in related variables and the spatial distribution in the spatial view. The clusters can be interactively refined based on the similarity analysis of biclusters. These steps are iteratively performed to verify the local relationships among variables, features, and scalar values.

\section{Results}
\label{sec:results}
In this section, three representative multivariate data sets in different domains were used to verify the effectiveness and usefulness of our framework in analyzing the local relationships among variables, biclusters, and scalar values. We performed all experiments on an Intel Core i7-7700K 4.20GHz CPU equipped with an NVIDIA GeForce GTX 1070 GPU. The minimum number of voxels for biclusters was set at 0.2\% of the total voxels of the explored volume to capture small features, such as the hurricane eye. In most simulations, biclusters corresponding to the background generally have a large number of voxels, and we filtered these less interesting biclusters based on the number of voxels (10\% of the total voxels).

\subsection{Hurricane Isabel Data Set}
\label{subsec:hurrican_isabel}

The hurricane Isabel data set has been widely used in previous research, and it presents a simulation of a hurricane created by the National Center for Atmospheric Research in the U.S. Ten variables were used in our experiment: PRE, PRECIP, QCLOUD, QGRAUP, QICE, QSNOW, QVAPOR, TC, and VEL (the magnitude of the wind speed). The resolution is $250 \times 250 \times 50$, and the 20th time step is chosen in our experiment to explore local relationships and classify the main features of the hurricane, i.e., the hurricane eye and the rainbands. The tolerance $\delta$ for generating the biclusters is 20 with coherent scalar-value patterns.

\begin{figure*}[]
 \centering
 \includegraphics[width=0.97\textwidth]{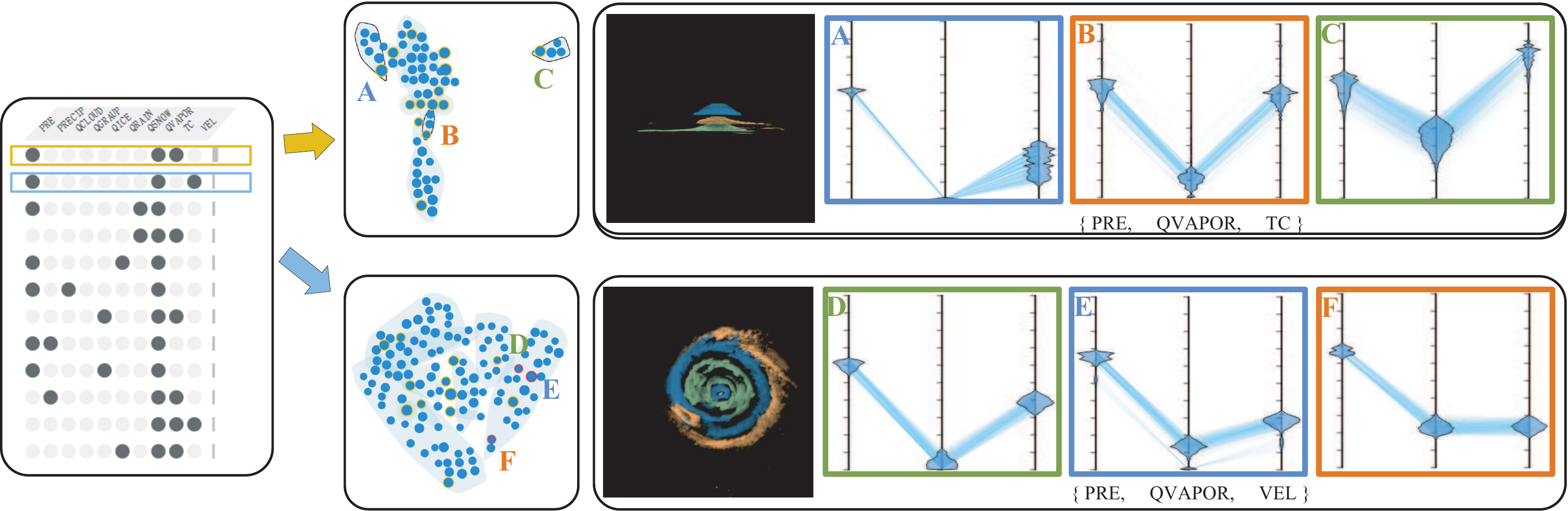}
 \caption{Visual exploration of the hurricane Isabel data set. Three groups in the first variable set labeled \emph{A, B,} and \emph{C} correspond to the upper (blue), middle (yellow), and lower (green) part of the hurricane eye. Their coherent scalar-value distributions are shown on the right of the spatial view. Three biclusters are selected in the second variable set \{PRE, QVAPOR, VEL\}. They are labeled \emph{D, E} and \emph{F} with red halos, which are the rainbands around the hurricane eye with an increasing distance (green-blue-yellow).}
 \label{fig:isable}
\end{figure*}

Because we are more interested in local relationships with at least three variables, we first filtered the variable set with only two variables in the association matrix. We sorted the variable sets according to the number of biclusters, which demonstrates the correlation of variables in terms of biclusters. As shown in Fig.~\ref{fig:isable}, the variable set \{PRE, QVAPOR, TC\} includes the most number of biclusters, and its biclusters are projected on the scatter plot to illustrate the similarity between them. Group \emph{C} on the right is far away from others, and the spatial view demonstrates that it is the lower part of the hurricane eye. When group \emph{A} on the left is selected, the upper part of the hurricane eye is presented in the spatial view. Based on these results, we can assume that the groups in the middle of the scatter plot may correspond to the middle part of the hurricane eye. Several groups are overlapped in the middle, and they represent nearly the same feature. To verify the assumption, we interactively merged these groups into one group \emph{B} and obtained the middle part of the hurricane eye. The scalar values of the three groups are displayed in the scalar-value views at the top of Fig.~\ref{fig:isable}.




The second variable set in the association matrix is \{PRE, QVAPOR, VEL\}.
The group on the right indicates the rainbands around the hurricane eye. When we further explored each bicluster in this group, there are three biclusters that correspond to three different rainbands near the hurricane eye. As shown in Fig.~\ref{fig:isable}, the rainbands are gradually away from the hurricane eye, and the scalar values of the pressure are almost the same. However, the water vapor mixing ratio is gradually increased, and the wind speed is gradually decreased. This also agrees with the knowledge of the rainbands around the hurricane eye.

Based on our co-analysis framework, we can quickly identify the variable set related to a local feature/phenomenon in multivariate data. By analyzing biclusters or their groups in the scalar-value view and spatial view, we can find that the variable set \{PRE, QVAPOR, TC\} is locally associated in the region of the hurricane eye, while the variable set \{PRE, QVAPOR, VEL\} is more useful in recognizing the rainbands around the hurricane eye.

\subsection{Turbulent Combustion Data Set}

This data set includes five variables: Heat Release Rate (HR), Mass Fraction of the Hydroxyl Radical (OH), Mixture Fraction (MIX), Scalar Dissipation Rate (CHI), and vorticity (VORT).

\begin{figure*}
 \centering
 \includegraphics[width=0.97\textwidth]{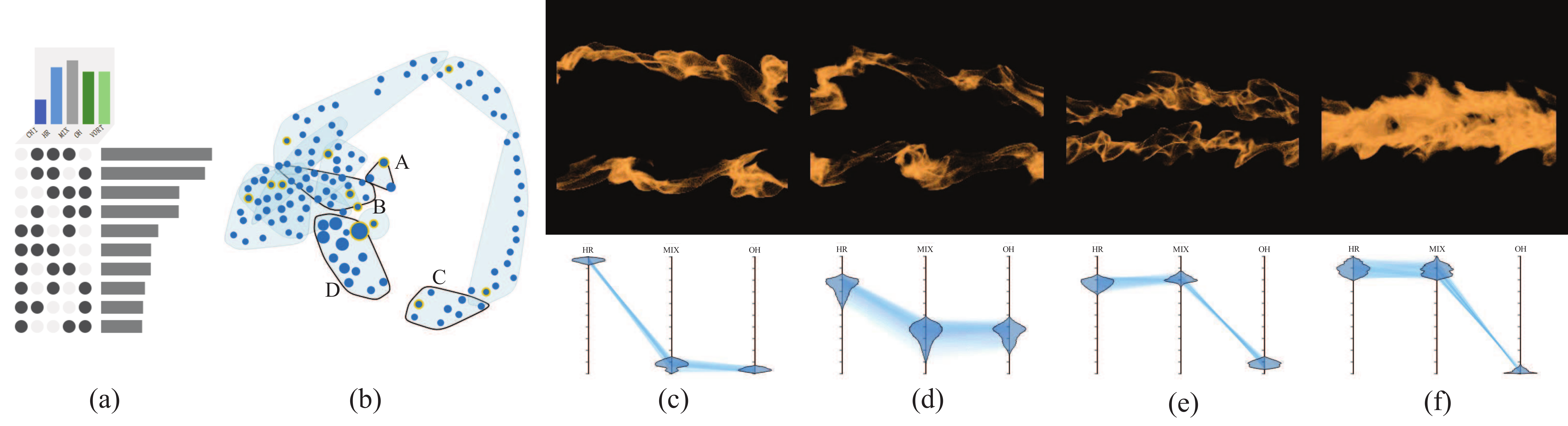}
 \caption{Visual exploration of the variable set \{HR, MIX, OH\} in the turbulent combustion data set with 65th time step and $240 \times 360 \times 60$ resolution. (a) The association matrix is sorted by the correlation value. (b) The biclusters of the variable set \{HR, MIX, OH\}. (c-f) The spatial and scalar-value distributions of groups \emph{A, B, C} and \emph{D}.}
 \label{fig:combustion}
\end{figure*}

We first sorted the variable sets with at least three variables by according to correlation of the variable set in the association matrix and selected the first variable set \{HR, MIX, OH\} to explore its biclusters as shown in Fig.~\ref{fig:combustion}(a). It is easy to identify four groups of biclusters that correspond to four parts of the flame in Fig.~\ref{fig:combustion}(b), i.e., the outer layer of the flame, the body of the flame, the inner layer of the flame, and the non-combustion region. The spatial distributions and scalar values of the four groups are illustrated in Fig.~\ref{fig:combustion}(c-f). It is evident that HR is high in the outer and inner layer of the flame and the non-combustion region, but OH is low, especially in the non-combustion region (nearly zero).

As shown in Fig.~\ref{fig:combustion}(a), the correlation value of the first variable set \{HR, MIX, OH\} is higher than the value of the second variable set \{HR, MIX, VORT\}. If we measure the correlation between variables based on all voxels~\cite{Biswas:2013:AIF}, instead of the voxels in biclusters, the most correlated variables with an informative variable MIX (the largest entropy) are OH and VORT, i.e., the third variable set. After exploring the third variable set, we find that its biclusters are less interested compared to the first variable set. Therefore, the correlation among variables can be better measured in the associated local regions, i.e., the regions of biclusters.


\subsection{Deep Water Impact Data Set}

This data set has been generated by a 3D simulation of a 250-meter-diameter asteroid impacting into ocean after passing through the atmosphere at an angle of $45^{\circ}$ in Los Alamos National Laboratory~\cite{Patchett:2018:OST}. Six variables were used for the experiment: pressure (prs), density in grams (rho), sound speed (snd), temperature (tev), volume fraction water (v02), and velocity, which is the magnitude of the wind speed.

\begin{figure*}[tb]
 \centering
 \includegraphics[width=0.97\textwidth]{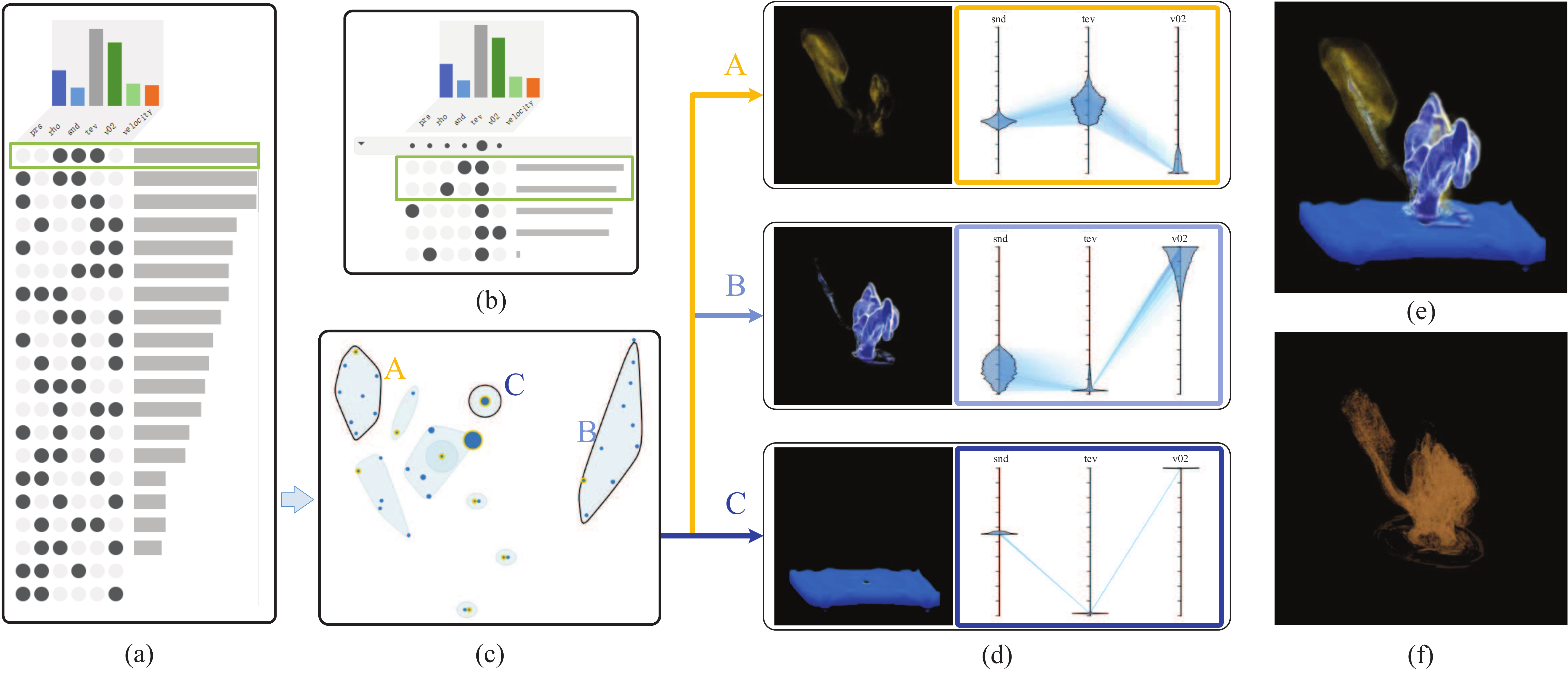}
 \caption{Visual exploration of the local relationships in the deep water impact data set with 27th time step and $150 \times 150 \times 150$ resolution. (a-b) The association matrix is sorted according to the number of biclusters and the variable set \{snd, tev, v02\} is on the top of the list. (c) Corresponding biclusters. (d) The spatial and scalar-value distributions of groups \emph{A} (a high temperature, high sound speed, and low water fraction in the air), \emph{B} (a low temperature, low sound speed, and high water fraction in the air), and \emph{C} (a low temperature, high sound speed, and high water fraction in the water). (e) The local features in groups \emph{A}, \emph{B} and \emph{C}. (f) The result of GSIM~\cite{Sauber:2006:MAA} for the correlation of the variable set \{snd, tev, v02\}.}
 \label{fig:asteroid}
\end{figure*}

Domain experts are interested in the effects of the phenomena on natural disasters such as rainfall. Rainfall is related to v02, which represents a fraction of water in the air or water vapor. Therefore, we selected the variable v02 as the starting variable to drill down to its children and further sorted the children according to the number of biclusters. As shown in Fig.~\ref{fig:asteroid}(b), tev and snd are primarily associated with v02. Alternatively, we can also sort the variable sets with at least three variables according to the number of biclusters as shown in Fig.~\ref{fig:asteroid}(a). The first variable set \{snd, tev, v02\} is also the one with the most number of biclusters, i.e., more local relationships. The biclusters of the variables set \{snd, tev, v02\} are projected on the scatter plot in Fig.~\ref{fig:asteroid}(c). There are several discernible groups, such as three distinguished groups \emph{A, B}, \emph{C}, and other groups have less interesting or coherent features. The spatial and scalar-value distributions of the three groups are displayed in Fig.~\ref{fig:asteroid}(d).


The region with a high temperature in group \emph{A} is primarily distributed around the asteroid's trajectory. The gravitational potential energy of the asteroid is converted into kinetic energy and the energy for overcoming air resistance. The energy for overcoming air resistance then turns into heat energy, increasing the temperature near the asteroid's trajectory. For group \emph{B}, it is easy to identify two regions with a high volume fraction of water (v02). One region is above the sea level impacted by the asteroid, and the other is the evacuated channel left by the asteroid's trajectory. For the former, the speed of the asteroid is reduced after impacting into the water, which causes the surrounding water to splash around and leads to an increase in the volume fraction of water above the impact position. A tsunami may occur when the impact is strong enough. For the latter, because of the high-temperature around the asteroid's trajectory, vast amounts of liquid water absorb heat and then change into water vapor, and the water molecules move and spread along the high temperature region. When there is enough water and sufficient suspended particles in a colder stratum, the water condenses together and produces rains if the water's gravity is higher than buoyancy. In addition, H$_{2}$O, a greenhouse gas, can absorb the reflected solar radiation of the Earth's surface so that it may increase the temperature around the area to some extent. Therefore, we conclude that there would be local rainfall with slight warming after the asteroid impacts into the ocean.


We compare our co-analysis framework with gradient similarity measure (GSIM)~\cite{Sauber:2006:MAA} using the variable set \{snd, tev, v02\}. Fig.~\ref{fig:asteroid}(e) presents the overall spatial distribution of groups \emph{A, B}, and \emph{C}. GSIM can measure the correlation of multiple variables by calculating the similarity among gradients at each voxel, and the result is presented in Fig.~\ref{fig:asteroid}(f). Overall, the spatial distributions are similar. However, our framework can effectively extract local features with similar scalar-value patterns, i.e., local relationships between variables and voxels, and each local feature includes a specific value combination revealing the local interaction of variables. In contrast, the result of GSIM is a global feature for the three variables, and it is challenging to gain insight into the local associations and their scalar-value distributions such as regions with high water vapor or low temperature.

\subsection{Discussion}

Our co-analysis framework provides a new perspective for systematically and visually exploring multivariate data. Three experiments demonstrate that our co-analysis framework can help users quickly explore variable sets of interest and discover the local correlations of the scalar values among different variables.


Compared to previous methods in the aspect of correlation analysis and multi-dimensional transfer functions, our framework clusters variables and voxels \emph{simultaneously} and extracts all biclusters with similar scalar-value patterns \emph{automatically}, and focuses on analyzing the \emph{local} relationships among variables, biclusters, and scalar values. In particular, our framework extends the analysis of value combinations of two variables~\cite{Liu:2016:AAF} to multiple variables such as three variables in our experiments. Although our system supports visual exploration of biclusters of all variables, we focus on the local correlations of 2-4 variables because the feature/phenomenon is generally associated with a subset of variables. Further, there have been many previous methods that use all variables to cluster features~\cite{Wu:2015:EAF}. As shown in Fig.~\ref{fig:asteroid}, our framework can effectively identify local features with similar scalar-value patterns from multiple variables, which is complementary to previous global correlation analysis~\cite{Sauber:2006:MAA}. Compared to interactive classification~\cite{Guo:2011:MTF}, our framework automatically generates all biclusters, groups biclusters to facilitate the exploration of biclusters, and designs coordinated views to identify variable sets of interest without much prior knowledge and to efficiently discover local correlations among variables.

Biclusters are generated in the preprocessing stage before the exploration process. Table~\ref{tab:performance} presents the computational performance of bicluster generation and the number of biclusters for three experimented data sets with different tolerances. The computational time ranges from less than 1 min to more than 1 h, and is roughly proportional to the number of biclusters, which depends on the number of variables and the complexity of the volume. The number of biclusters can be very large, especially for 10 variables such as the hurricane data. During visual exploration, users can filter small or large biclusters based on the number of voxels according to their analysis requirements, including the background feature. Because we assume that users have little prior knowledge about the data, biclusters in each combination of variables are extracted from the data. If users have prior knowledge, only variables of interest need to search for biclusters and this would greatly improve computational and analytical efficiency. In addition, the searching process can be accelerated by parallel computation because the expansion of each variable pair is independent.

\begin{table}
    \centering
     \caption{The computational time (seconds) of bicluster generation and the number of biclusters for three data sets.}
     \label{tab:performance}
    \begin{tabular}{|c|c|c|c|c|c|c|c|}

     \hline
     \multirow{2}*{$\delta$} & \multicolumn{2}{c|}{Isabel} & \multicolumn{2}{c|}{Combustion} & \multicolumn{2}{c|}{Asteroid} \\ \cline{2-3} \cline{4-5} \cline{6-7}
                        &  time  &   number  &  time  &   number   &  time  &   number    \\
     \hline
 	10 & 4924 & 6159 & 451 & 904 & 131 & 985 \\
	 15 & 2423 & 4352 & 460 & 1024 & 72 & 810  \\
      20 & 1070 & 3040 & 472 & 923 & 64 & 756  \\
	 25 & 996 & 2370 & 482 & 923 & 60 & 636  \\
      30 & 526 & 1896 & 212 & 650 & 40 & 539  \\
     \hline
    \end{tabular}
\end{table}

The biclustering method includes two parameters, the tolerance and the minimal number of voxels. The minimal number of voxels can be set to a very small number even zero to determine all biclusters. However, this would increase the computational time and generate many noisy and meaningless biclusters with only a few voxels. In our experiments, the minimal number of voxels is fixed to 0.2\% of the total voxels of the explored volume, a relatively low value, to balance these factors. In this case, the resolution of multivariable data has little impact on the number of biclusters, although small biclusters may not be extracted when the resolution is low. The tolerance is related to the data range and meaning of the variables, and determines the maximum difference of scalar values to be considered as the same feature. As shown in Table~\ref{tab:performance}, the number of biclusters decreases with an increase in the tolerance because many small biclusters are merged into large biclusters. In order to restrain the over-segmentation problem, the tolerance can be chosen from 10 to 30 depending on the application (20 by default in our experiments).


A potential limitation of biclustering is that it would generate too many biclusters from multivariate data, which results in a long analysis process. Our framework can group biclusters to reduce the number of biclusters to be analyzed, and supports filtering based on the number of voxels for removing too small or too large biclusters, including noisy and background features. As demonstrated in the experiments, the association matrix can sort the variable sets based on the number of biclusters and the correlation of the variable set for exploring interesting variable sets and biclusters. In the future, this issue can be further addressed by recommending meaningful groups or biclusters in different variable sets by automatically evaluating the coherence of scalar values and the spatial distribution. In addition, the variables of the bicluster with a similar scalar-value pattern may be mathematically correlated, but they may not be correlated in the physical phenomena. Therefore, this requires users to further verify such correlation in the bicluster with domain knowledge and human intelligence during visual exploration.

\section{Conclusion}

In this paper, we proposed a co-analysis framework to guide the visual exploration of the local correlations in multivariate data based on biclusters. The biclustering method is used to automatically generate all biclusters only containing voxels with a similar scalar-value pattern over multiple variables. They are organized and grouped hierarchically to reduce the complexity of user interaction and are visually presented in four coordinated views to facilitate interactive exploration of multivariate data from different facets of multivariate data. Experiments demonstrated that our co-analysis framework can effectively identify the associated variable set related to a local feature/phenomenon, compare the similarity of biclusters, and analyze the correlations of the scalar values of different variables in local regions.

For future work, we plan to recommend meaningful groups or biclusters in different variable sets to further improve analytical efficiency, and employ parallel computation to accelerate computational efficiency in bicluster generation. We would also like to extend our co-analysis framework to time-varying multivariate data for capturing the coherence in the time space.


\section*{Acknowledgement}{
This work was supported by the National Key Research \& Development Program of China (2017YFB0202203), National Natural Science Foundation of China (61472354 and 61672452), NSFC-Guangdong Joint Fund (U1611263).
}

\section*{References}

\bibliography{mybibfile}

\end{document}